\documentclass[sigconf]{acmart}

\usepackage{graphicx}
\usepackage{amsmath}
\usepackage{booktabs}
\usepackage{times}
\usepackage{epsfig}
\usepackage{graphicx}
\graphicspath{{images/}}
\usepackage{subcaption}
\usepackage{amsthm}
\usepackage{amsfonts}
\usepackage{multirow}
\usepackage{microtype}
\usepackage{tabularx}
\usepackage{xspace}
\usepackage{caption} 
\usepackage{enumitem}
\usepackage{hyperref}

\AtBeginDocument{%
  \providecommand\BibTeX{{%
    \normalfont B\kern-0.5em{\scshape i\kern-0.25em b}\kern-0.8em\TeX}}}

\setcopyright{acmcopyright}
\copyrightyear{2022} 
\acmYear{2022} 
\setcopyright{acmcopyright}\acmConference[MM '22]{Proceedings of the 30th ACM International Conference on Multimedia}{October 10--14, 2022}{Lisboa, Portugal}
\acmBooktitle{Proceedings of the 30th ACM International Conference on Multimedia (MM '22), October 10--14, 2022, Lisboa, Portugal}
\acmPrice{15.00}
\acmDOI{10.1145/3503161.3548409}
\acmISBN{978-1-4503-9203-7/22/10}

\newcommand{\name}{ReFormer\xspace}

\usepackage[normalem]{ulem}

\newcommand{\del}[1]{}  
\newcommand{\note}[1]{}
\newcommand{\rev}[1]{\textcolor{black}{{#1}}}

\begin{document}

\title{ReFormer: The Relational Transformer for Image Captioning}

\author{Xuewen Yang}
\authornote{The author is currently a senior research scientist at InnoPeak Technology, Inc. He was previously a PhD student in Stony Brook University, using email xuewen.yang@stonybrook.edu.}
\email{xuewen.yang@innopeaktech.com}
\orcid{0000-0002-9782-0499}
\affiliation{%
  \institution{InnoPeak Technology, Inc.}
  \streetaddress{2479 E Bayshore Rd \#110}
  \city{Palo Alto}
  \state{CA}
  \country{USA}
}

\author{Yingru Liu}
\email{liu2231665@hotmail.com}
\affiliation{%
  \institution{Stony Brook University}
  \city{Stony Brook}
  \state{NY}
  \country{USA}}

\author{Xin Wang}
\email{x.wang@stonybrook.edu}
\affiliation{%
  \institution{Stony Brook University}
  \city{Stony Brook}
  \state{NY}
  \country{USA}
}


\begin{abstract}
Image captioning is shown to be able to achieve a better performance by using scene graphs to represent the relations of objects in the image. The current captioning encoders generally use a Graph Convolutional Net (GCN) to represent the relation information and merge it with the object region features via concatenation or convolution to get the final input for sentence decoding. However, the GCN-based encoders in the existing methods are less effective for captioning due to two reasons. First, using the image captioning as the objective (i.e., Maximum Likelihood Estimation) rather than a relation-centric loss cannot fully explore the potential of the encoder. Second, using a pre-trained model instead of the encoder itself to extract the relationships is not flexible and cannot contribute to the explainability of the model. To improve the quality of image captioning, we propose a novel architecture \name - a RElational transFORMER to generate features with relation information embedded and to explicitly express the pair-wise relationships between objects in the image. \name incorporates the objective of scene graph generation with that of image captioning using one modified Transformer model. This design allows \name to generate not only better image captions with the benefit of extracting strong relational image features, but also scene graphs to explicitly describe the pair-wise relationships. Experiments on publicly available datasets show that our model significantly outperforms state-of-the-art methods on image captioning and scene graph generation.
\end{abstract}




\maketitle

\section{Introduction}
Research on image captioning to generate textual descriptions of images has made a great progress in recent years thanks to the introduction of encoder-decoder architectures~\cite{Anderson2018,Aneja18,Johnson16,Karpathy2017,Lu2018NeuralBT,Venugopalan2017,Kelvin2015,Yang2020Fashion}.
Existing models are generally trained and evaluated on datasets created for image captioning  like COCO~\cite{ChenCOCO15,Lin14} and Flickr~\cite{Hodosh13} that only contain generic object categories but not pair-wise relations of the objects in the image.

To equip the captioning model with relation information, some more recent studies resort to the scene graph generation~\cite{xu2017scenegraph,zellers2018scenegraphs} to provide the graph representations of real-world images with the semantic summaries of objects and their pair-wise relationships. 
For example, the graph in Figure~\ref{fig:scene-graph} encodes the  key objects in the image such as people (`man'), their possessions (`hair' and `shirt', both possessed by the man), and their activities (the man is `holding' a `racket'). 
The graph representation has been applied to improve the image related tasks that involve natural language~\cite{Teney17,yin2017obj2text}.
When it comes to the task of image captioning, recent studies~\cite{wang2019role,yang2019,Yao2018} propose to first use a scene graph generation model well-trained on Visual Genome~\cite{Krishna17} dataset to predict the pair-wise relationships existing in the COCO image  and then use a Graph Convolutional Net (GCN) to encode the relation information. 
Typically, the object region features and the relation representations are then merged together via concatenation or convolution to feed into a decoder for generating a sentence using the Maximum Likelihood Estimation (MLE).
These methods typically suffer from at least one of three main weaknesses:
(i) There are mis-alignments between the image objects and the relation labels, because the regions containing the objects do not correspond to those used to predict the relations;
(ii) Given that the goal of using a GCN is to extract the relation information,  \rev{the training of model for GCN is less effective by only using the objective to optimize the captioning} without considering the object relationship;
(iii) The encoder itself cannot extract the relations between objects but relying on other pre-trained models to do it, which makes the captioning less explainable.
As another observation, recent studies~\cite{agrawal2016analyzing,caglayan2019,devlin2015exploring,Goyal2019,shekhar2019} have pointed out that good metric scores can be achieved with a strong decoder, without the need of underlying encoder to truly understand the visual content.
Thus, it becomes less likely to determine if  the models are really learning some important relationships through the encoder or they just follow some language rules by the decoder.
To be more concrete, for a generated sentence like `a man is riding a bike', can the model really tell the difference among `riding', `rolling' or `on' or it just follows some language expression rules (\textit{i.e.}, `riding' is more commonly used than `rolling' and `on')?

Regularizing the encoder with a relation-centric objective is essential since it can not only \rev{guide} the encoder to learn representations with relation information embedded, but also explicitly express the pair-wise relationships and explain the generation of some relational words.
In this paper, we propose \textit{\name}: the RElational transFORMER that learns a scene graph to express the object relationships in the process of decoding a sentence description. 
\name incorporates both image captioning and scene graph generation components via a novel transformer encoder.
Different from conventional Transformer~\cite{Vaswani2017} that only uses the image captioning as the final objective to train both the encoder and the decoder, \name uses a scene graph generation objective to \rev{guide} the encoder to learn better relational representations.
Since the image captioning and scene graph generation are two distinct tasks, directly using the Multi-Task Learning paradigm is non-trivial.
We propose a \textit{sequential training} algorithm that \rev{guides} the \name to learn both tasks step by step.

Our work has three main contributions.
(i) We propose to generate scene graphs as a way to enrich the captions that they together can better describe the images;
(ii) We design a novel relational Transformer (\name) that can better learn the image features for captioning with the relationships embedded via an auxiliary scene graph generation task;
(iii) We propose a \textit{sequential training} algorithm that \rev{guides} the \name to accomplish both tasks in three consecutive steps.
Experimental results show that \name can achieve better performance than state-of-the-art methods on both image caption generation and scene graph generation.
We will release the source code.

\section{Background and Related Work}
In this section, we first introduce the background knowledge of scene graph generation, and then discuss the related work on image captioning and the application of scene graphs in image captioning.
\subsection{Scene Graph Generation}
\label{sec:sgg}
\begin{figure}
    \centering
    \small
    \includegraphics[width=0.45\textwidth]{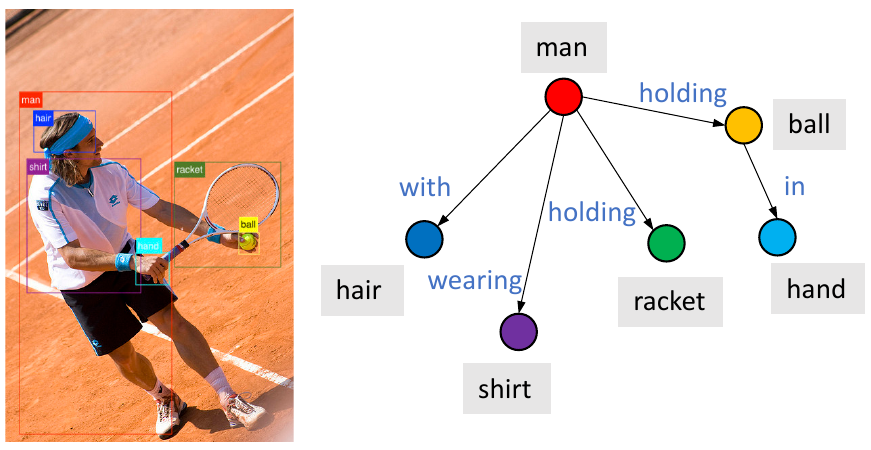}
    \caption{A scene graph containing entities, such as \texttt{man}, \texttt{hair} or \texttt{racket}, that are localized in the image with bounding boxes and the relationships between those entities, such as \texttt{with}, \texttt{wearing} and \texttt{holding}.}
    \label{fig:scene-graph}
\end{figure}
A \textit{scene graph}, $G$, as shown in Figure~\ref{fig:scene-graph}, is a structural representation of the semantic contents in an image~\cite{Krishna17}.
It consists of:
\begin{itemize}[noitemsep]
    \item a set of \textit{bounding boxes} $B=\{b_1,\ldots,b_m \}$, $b_i=(x_{i1},y_{i1},x_{i2},y_{i2})$\footnote{$x_{i1},y_{i1}$ are the top-left coordinates of $b_i$, while $x_{i2},y_{i2}$ are the bottom-right coordinates.}.
    \item a corresponding set of \textit{objects} $O=\{o_1,\ldots,o_m \}$, where  $o_i\in \mathcal{C}$ is the class label assigned to the bounding box $b_i$ and $\mathcal{C}$ is the set containing all label categories.
    \item a set of pair-wise relationships $R=\{\ldots, r_{i\rightarrow j},\ldots \}$, with $r_{i\rightarrow j}\in \mathcal{R}$ representing the relationship between a start node $(b_i,o_i)$ and an end node $(b_j,o_j)$. $\mathcal{R}$ is the set of relation types, including the `background' predicate, which indicates that there is no edge between the specified objects.
\end{itemize}

Scene graph~\cite{tang2020unbiased,tang2018learning,zellers2018scenegraphs} is often generated with a few procedures: object detection (detecting $b_i$), classification (classifying $o_i$) and predicate (relation label) prediction to determine $r_{i\rightarrow j}$ given $(b_i,o_i)$ and $(b_j,o_j)$.
Most of the methods on scene graph generation have been developed on the Visual Genome~\cite{Krishna17} dataset, which provides annotated scene graphs for $100$K images, consisting of over $1$M instances of objects and $600$K relations.
Since only a small portion of images in this data set also exist in the COCO captioning dataset~\cite{ChenCOCO15}, directly using a multi-task learning scheme on both tasks is challenging.

\subsection{Image Captioning}
State-of-the-art approaches~\cite{Anderson2018,He2020image,Johnson16,Sammani_2020_CVPR,wang2020unique,Kelvin2015,Yang2020Fashion,yingru,yingruliu2,xuewen_mm20,feng2015} 
mainly use encoder-decoder frameworks with attention to generate captions for images.
Xu \textit{et al.}~\cite{Kelvin2015,yang2018cross,yang2019latent,yang-etal-2021-journalistic} developed soft and hard attention mechanisms to focus on different regions in the image when generating different words.
Similarly, Anderson \textit{et al.}~\cite{Anderson2018} used a Faster R-CNN~\cite{Ren15} to extract regions of interest that can be attended to.
Yang \textit{et al.}~\cite{Yang2020Fashion} used self-critical sequence training for image captioning. 

Various Transformer-based~\cite{Vaswani2017} models have achieved promising success on the
image captioning task~\cite{cornia2020m2,He2020image,Simao2019,Li_2019_ICCV}.
Cornia \textit{et al.}~\cite{cornia2020m2} proposed a meshed-memory transformer that learns a multi-level representation of the image regions, and uses a mesh-like connectivity at decoding stage to exploit low- and high-level features.
Li \textit{et al.}~\cite{Li_2019_ICCV} introduced the entangled attention that enables the Transformer to exploit semantic and visual information simultaneously.
He \textit{et al.}~\cite{He2020image} introduced the image transformer, which consists of a modified encoding transformer and an implicit decoding transformer to adapt to the structure of images.
Herdade \textit{et al.}~\cite{Simao2019} introduced the object transformer, that explicitly incorporates information
about the spatial relationship between detected objects through geometric
attention.

Some research studies have been using scene graphs for image captioning~\cite{guo2019vsua,wang2019role,Yao2018,zhong2020comprehensive}.
Zhong \textit{et al.}~\cite{zhong2020comprehensive} proposes a scene graph decomposition method that decomposes a scene graph into a set of sub-graphs, with each sub-graph capturing a semantic component of the input image.
By selecting important sub-graphs, different target sentences are decoded.
Wang \textit{et al.}~\cite{wang2019role} uses two encoders, one is a ResNet image encoder, the other is a GCN encoder for the relation labels of the objects.
The two encoders are then attached with an LSTM decoder and combined together using attention.
Yao \textit{et al.}~\cite{Yao2018} proposes to use two GCNs to encode the spatial and semantic relations in an image.
Then the features from the two encoders are merged together via attention to get the final feature.
Guo \textit{et al.}~\cite{guo2019vsua}
proposed to explicitly model the object interactions in semantics and geometry based on Graph Convolutional Networks (GCNs). 
Yang \textit{et al.}~\cite{yang2019} proposed the scene graph auto-encoding technique to learn a dictionary that helps to encode the desired language prior, which guides the encoding-decoding pipeline.

\name differs from the previous methods in 
two aspects: (i) \name can not only generate a caption but also a scene graph to capture the relationships between the objects without using an external scene graph generator.
(ii) \name integrates the scene graph generation with image captioning using a \textit{sequential training} algorithm to better learn the relational image features step by step.

\section{Design of \name}
In this section, we first discuss the problems of existing schemes and propose the  basic architecture to construct the \name that combines the scene graph generation with image captioning to learn relational features of images.
We then present the scene graph generation task which is used to first pre-train the encoder of \name and later used as an auxiliary objective to generate captions. 
Finally, we describe the image captioning task as well as the training algorithm to train on both tasks.

\subsection{A Relational Encoding Learning Idea}
\label{sec:intuition}
In a standard  image captioning model based on encoder-decoder structure, the decoder directly predicts the target sequence $\mathbf{y}$ conditioned on the source input $\mathbf{x}$. 
The captioning probability $P(\mathbf{y}\vert \mathbf{x})$ is modeled directly using the probability of each target word $\mathbf{y}_i$ at the time step $i$ conditioned on the source input sequence $\mathbf{x}$ and the current partial target sequence $\mathbf{y}_{1:i-1}$ as follows:
\begin{equation}
\small
P(\mathbf{y} \vert \mathbf{x}; \boldsymbol{\theta}) = \prod_{i=1}^{N} P(\mathbf{y}_i\vert \mathbf{x}, \mathbf{y}_{1:i-1};\boldsymbol{\theta})
\label{eq:img_cap}
\end{equation}
where $\boldsymbol{\theta}$ denotes the parameters of the model\footnote{Through-out this paper, we omit $\boldsymbol{\theta}$ for simplicity.}.

\begin{figure}
\centering
\small
\begin{subfigure}[b]{0.45\textwidth}
   \includegraphics[width=1\linewidth]{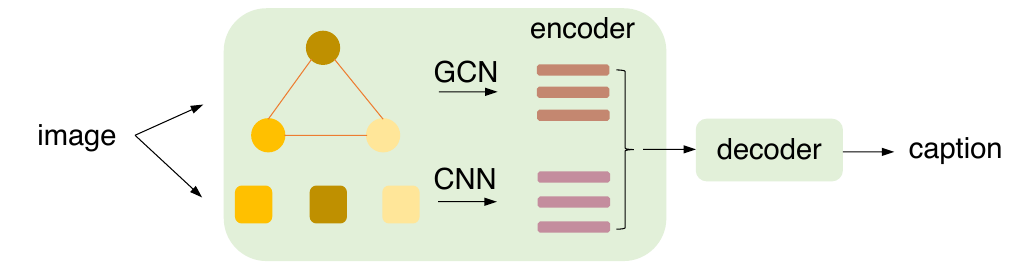}
   \caption{Two-encoder captioning model.}
   \label{fig:gcn} 
\end{subfigure}

\begin{subfigure}[b]{0.45\textwidth}
   \includegraphics[width=1\linewidth]{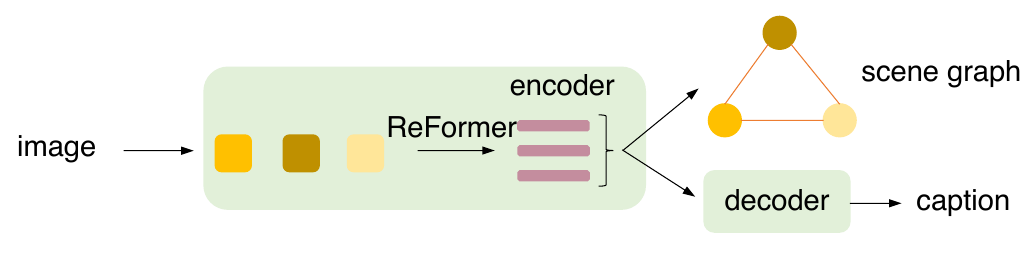}
   \caption{\name: Our Relational Transformer model.}
   \label{fig:reformer}
\end{subfigure}

\caption[]{\small{Models trying to encode relation information: (a) A widely-used two-encoder captioning model that one encoder is a GCN used to encode the scene graphs into relational features and the other is a pre-trained CNN (\textit{i.e.} ResNet) for images.
(b) \name, our Relational Transformer that generates a scene graph and a corresponding caption.}}
\end{figure}

In general, $\mathbf{x}$ are image features that can be obtained by feeding an image $X$ to a pre-trained CNN (\textit{e.g.} ResNet) encoder $\mathbf{x}=CNN(X)$.
To integrate relational information into the image features, some state-of-the-art methods first use a well-trained scene graph generation model to extract a graph $g$ from the same image and then use a Graph Convolutional Net (GCN) to encode $g$ to vectors, shown in Figure~\ref{fig:gcn}.
Then the image features and the relational features are merged together $\mathbf{x}=[CNN(X);GCN(g)]$, where $[\cdot;\cdot]$ can be concatenation, attention or convolution.
This straightforward way of integrating relational information may suffer from a few problems.
First, training a GCN with the objective of getting the image caption (Maximum Likelihood Estimation, \textit{i.e.}, MLE) might be less effective.
The goal of using a GCN is to capture the relational information existing in the image,
while the likelihood function used to estimate the probability distribution is not directly relevant to the relation of objects in an image.
Second, if the model no longer has access to the pre-trained scene graph generator (\textit{e.g.}, using a different dataset), the caption generation might not be feasible any more.
Third, recent studies show that good metric scores can be obtained with a strong decoder, without the underlying encoder to truly understand the visual content. This indicates that the caption objective alone cannot effectively guide the training of the encoder to accurately extract the relationships between the objects. 

The above analysis motivates us to design a model that can generate scene graphs together with the learning of image caption using the same dataset  without the need of an external scene graph generator, but with the objective of producing the good scene graph. 
In this way, we can also better train the encoder to well understand the visual contents without being misled by the good results from a stronger decoder.
We propose a second objective $P(g \vert X)$ for good  scene graph generation and apply it to the encoder learning.
We use a Transformer model where the encoder is used to generate scene graphs and the decoder is applied to generate captions.
By incorporating this `RElational' objective, our transFORMER can also better learn the image features with relation information embedded. 
The simplified model structure of \name is shown in Figure~\ref{fig:reformer}.


\begin{figure*}
    \centering
    \includegraphics[width=0.9\textwidth]{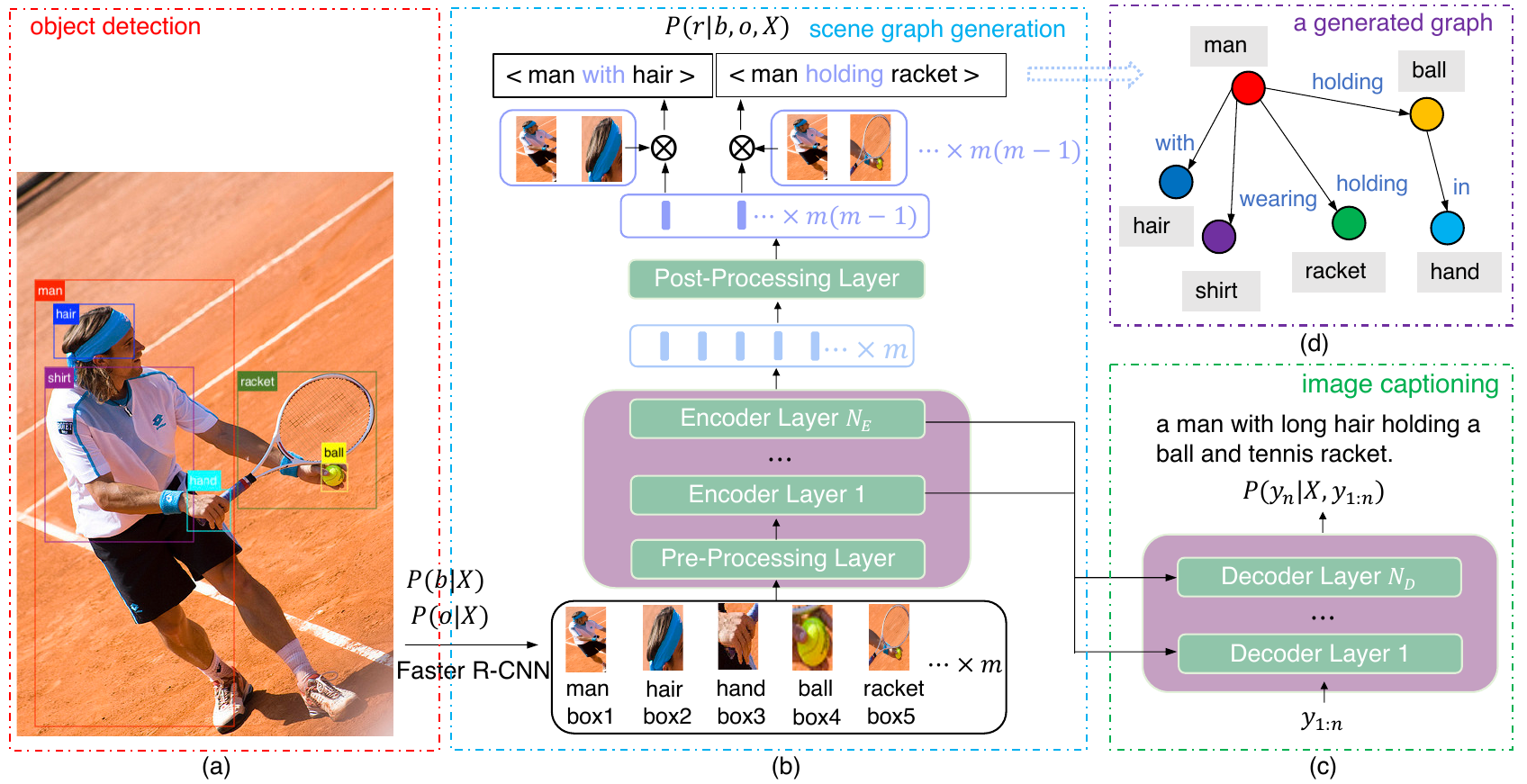}
    \caption{\small{The architecture of \name. \name consists of a (a) Faster R-CNN object detection model to provide $m$ bounding boxes, object labels and object region features, a (b) Transformer encoder to generate $m\times (m-1)$ pair-wise relations for the objects in the image and a (c) Transformer decoder for generating captions. The generated scene graph is shown in (d).}}
    \label{fig:model}
\end{figure*}

\subsection{Scene Graph Generation}
\label{sec:sggpt}

We first lay out the mathematical formulation of the scene graph generation problem.
As introduced in Section~\ref{sec:sgg}, for an image of $m$ objects, its visually grounded scene graph consists of tuples $\big(r_{i\rightarrow j}, (b_i, o_i), (b_j, o_j)\big)_{i,j=1,i\neq j}^m$\footnote{$i\in {1\ldots m}$, $j\in {1\ldots m}$, but $i\neq j$.}, with $b_i$ being the bounding box of the object $i$, $o_i$ the label and $r_{i\rightarrow j}$ the predicate (relation label) between objects $i$ and $j$.
Thus, the scene graph generation objective can be derived as follows:
\begin{equation}
\footnotesize
\begin{aligned}
    P(g \vert X)
    =& \prod_{(i,j)}^{m} P(r_{i\rightarrow j}\vert (b_i,o_i),(b_j,o_j),X) \\
    &\times \prod_{k}^{m}P(o_k\vert b_k,X)P(b_k\vert X)
\end{aligned}
\label{eq:sgg_obj}
\end{equation}
To simplify the learning process and improve the effectiveness of training, we conveniently use Faster-RCNN to model $P(o_k\vert b_k, X)$ and $P(b_k\vert X)$ together.
The negative log-likelihood of the model's parameters computed on the training data is then given by:
\begin{equation}
\footnotesize
\begin{aligned}
    \mathcal{L}_g&=-\log P(g \vert X) 
    = -\sum_{(i,j)}^{m} \log P(r_{i\rightarrow j}\vert (b_i,o_i),(b_j,o_j),X)\\ 
    &- \sum_{k}^{m}[\log P(b_k\vert X) + \log P(o_k\vert b_k, X)]
\end{aligned}
\label{eq:sgg_llh}
\end{equation}
We model $P(b_k\vert X)$ and $P(o_k\vert b_k, X)$ with object detector Faster R-CNN~\cite{Ren15} to automatically generate a set of bounding boxes $b_k$ and the corresponding object labels $o_k$ from an image $X$ (Figure~\ref{fig:model}(a)).
In practice, training a Faster R-CNN is highly non-trivial. To simplify the training of our proposed \name, we first train  a Faster R-CNN till its convergence. Then with its parameters learnt, we train the model $P(r_{i\rightarrow j}\vert (b_i,o_i), (b_j, o_j),X)$ (Figure~\ref{fig:model}(b)) with $(b_i, o_i)$, $(b_j,o_j)$ and $X$ as the input.
In theory, if the groundtruth $b_k$ and $o_k$ are provided, we can eliminate Faster R-CNN. 
To feed in $b_k$, $o_k$ and $X$, we use a Pre-Processing Layer (Section~\ref{sec:ppl}) which converts the three inputs into vectors and then concatenate them together to get the final input features.



Modeling $P(r_{i\rightarrow j}\vert(b_i,o_i),(b_j,o_j),X)$ is actually a predicate classification task.
We use the \name encoder (Section~\ref{sec:el}) to first encode the input into high-level features and then apply a Post-Processing Layer (Section~\ref{sec:popl}) to get the final output for classification.
The features right before the Post-Procesing layer contain useful relational information and are used as the input to the decoder for caption generation (Section~\ref{sec:ic}).
We talk about more details about the model architecture in Section~\ref{sec:re}.

\subsection{Encoder Architecture}
\label{sec:re}

The encoder takes as input a set of $m$ image region tuples,
$(b_i, o_i, \mathbf{v}_i)_{i=1}^m$, where $\mathbf{v}_i$ is defined as the mean-pooled convolutional feature from region $i$ with dimension $2048$ and the number of region varies for different images.
It consists of three main components which process the input consecutively:
(i) a Pre-Processing Layer that takes the bounding boxes, object labels and image region features as input and linearizes them to form the input vector; (ii) the Encoder Layers further process the features  with Multi-Head Attention to create a contextualized representation of each object; and (iii) the Post-Processing Layer where the object features are paired to make predictions of their relationships.

\subsubsection{Pre-Processing Layer}
\label{sec:ppl}
Different from conventional image captioning model where only image region features $\mathbf{v}_i$ are used as the input, we also use object labels $o_i$ as well as bounding boxes $b_i$ as the input to encode both object information and spatial relation information.

Given the bounding box $b=(x_1, y_1, x_2, y_2)$, to better represent its location as well as size in the image, we normalize it with the size of the image and convert it into a $9$-dimensional vector $(\frac{c_x}{W}, \frac{c_y}{H}, \frac{w}{W}, \frac{h}{H}, \frac{x_1}{W}, \frac{y_1}{H}, \frac{x_2}{W}, \frac{y_2}{H}, \frac{wh}{WH})$, where$(c_x, c_y)$ is the coordinate of the bounding box center, $(w, h)$ and $(W, H)$ are the width and height of the bounding box and the image respectively. 
We use a two-layer feed-forward net to encode it into vector $\mathbf{v}_b$ of dimension $d_b$.

As object labels (\textit{e.g.} `man') are meaningful words from natural languages, we use Glove~\cite{pennington2014glove} features, which contain pre-trained features to capture fine-grained semantic and syntactic regularities in natural languages.
The dimension of the object label feature $\mathbf{v}_l$ is denoted as $d_l$.

Thus, the final image feature is represented as $\mathbf{v}=f([\mathbf{v}_i; \mathbf{v}_b; \mathbf{v}_l])$, where $[\cdot;\cdot;\cdot]$ is the concatenation operation, and $f(\cdot)$ is a feed-forward network to map the feature to dimension $d_h$.

\subsubsection{Transformer Encoder Layer}
\label{sec:el}
Given a set of image features $\mathbf{v}=\{\mathbf{v}_1,\ldots, \mathbf{v}_m\}$ extracted using the Pre-Processing Layer, we use
a stack of $N_E$ Transformer~\cite{Vaswani2017} encoder layers to obtain a permutation invariant encoding.
Each encoder layer consists of a multi-head self-attention layer followed by a small feed-forward network. 
The multi-head self-attention layer itself consists of $h$ identical heads. 
Each attention head first calculates the queries $\mathbf{Q}$, keys $\mathbf{K}$ and values $\mathbf{V}$ for the $m$ input features as follows:
\begin{equation}
\small
    \mathbf{Q}=\mathbf{X}\mathbf{W}_Q,\mathbf{K}=\mathbf{X}\mathbf{W}_K,\mathbf{V}=\mathbf{X}\mathbf{W}_V,
    \label{eq:qkv}
\end{equation}
where $\mathbf{X}$ contains all the input vectors $\mathbf{x}_1,\ldots,\mathbf{x}_m$ stacked into a matrix and $\mathbf{W}_Q$, $\mathbf{W}_K$, $\mathbf{W}_V$ are learned projection matrices.
The image features are used as the input to the first self-attentive layer.
For the following layers, we use the output of the previous encoder layer as the input to the current layer.
The output of each head are then computed via scaled dot-product attention without using any recurrence:
\begin{equation}
\small
    head(\mathbf{X})=Attention(\mathbf{Q},\mathbf{K},\mathbf{V})=softmax(\frac{\mathbf{Q}\mathbf{K}^{T}}{\sqrt{d}})\mathbf{V}
    \label{eq:att}
\end{equation}
where $d$ is a scaling factor.
Eq.~\ref{eq:qkv} and \ref{eq:att} are calculated for every head independently. 
The output of all $h$ heads are then concatenated to one output vector and multiplied with a learned projection matrix $\mathbf{W}_O$:
\begin{equation}
\small
    MultiHead(\mathbf{X})=Concat(head_1,\ldots,head_h)\mathbf{W}_O
    \label{eq:mha}
\end{equation}
The multihead attention is then fed into a point-wise feed-forward network (FFN), which is
applied to each output of the attention layer:
\begin{equation}
\small
    \mathbf{h}=FFN(\mathbf{x})=\max(0, \mathbf{x}\mathbf{W}_1+\mathbf{b}_1)\mathbf{W}_2+\mathbf{b}_2
\end{equation}
where $\mathbf{W}_1$, $\mathbf{b}_1$ and $\mathbf{W}_2$, $\mathbf{b}_2$ are the weights and biases of two fully connected layers. 
In addition, skip-connections and layer-norm are applied to the outputs of the self-attention and the feed-forward
layer.


Other choices like LSTM or convolutional layers are also feasible.
In this paper, we choose self-attention layers because self-attention operation can be seen as a way of encoding pair-wise relationships between input features given that the self attention weights depend on the pair-wise similarities between the input features.
This is coherent to the objective of the paper: to model the relationships from the input objects.

\subsubsection{Post-Processing Layer}
\label{sec:popl}
For an image of $m$ objects, there are $m(m-1)$ possible relationships.
For each possible relationship between object $i$ and $j$, we compute the probability of the relationship of label $r_{i\rightarrow j}$.
Since the output of the last Transformer encoder layer $\mathbf{h}$ contains $m$ features $\mathbf{h}_1,\ldots,\mathbf{h}_m$, directly predicting the relationships using these features is impossible.
We thus use a Post-Processing Layer that maps $m$ object features into $m(m-1)$ pair-wise features.
We use a Linear layer to map $\mathbf{h}_i$ of dimension $d_h$ into $\mathbf{r}_i$ of dimension $2d_h$.
By doubling the dimension of $\mathbf{h}_i$, we can then equally split $\mathbf{r}_i$ into two parts, head $\mathbf{r}_i^h$ and tail $\mathbf{r}_i^t$, with head standing for the relationship starts from this node while tail stands for the relationship ends at this node.
Thus, for each pair-wise relationship representation, we can have:
\begin{equation}
\small
    \mathbf{r}_{i\rightarrow j}=(\mathbf{W}_r[\mathbf{r}_i^h;\mathbf{r}_j^t])\odot[\mathbf{v}_i;\mathbf{v}_j]
\end{equation}
where $\odot$ is point-wise multiplication operation, $[\cdot ; \cdot]$ is concatenation and $\mathbf{W}_r$ is a trainable parameter.
The distribution is:
\begin{equation}
\small
    P(r_{i\rightarrow j}\vert (b_i,o_i), (b_j,o_j),X)=softmax(\mathbf{r}_{i\rightarrow j})
\end{equation}

\subsection{Weighted Decoder for Image Captioning}
\label{sec:ic}
As discussed in Section~\ref{sec:intuition},
the objective of our model for caption generation is to minimize the negative log-likelihood of the correct caption using the maximum likelihood estimation: 
\begin{equation}
\small
    \mathcal{L}_{c} = -\sum_{i=1}^{n}\log p(\mathbf{y}_i\vert \mathbf{y}_{1:n-1}, \mathbf{x})
\label{eq:loglikelihood}
\end{equation}
The decoder that is used to model Eq.~\ref{eq:loglikelihood} consists of a stack of $N_D$ decoder layers (Figure~\ref{fig:model}(c)).
For each layer, to calculate the distribution for the word at the time step $i$, it takes as input: the embeddings of all previously generated words $\mathbf{y}_{0:i-1}$ and the context embeddings $\mathbf{x}$ from the encoder.

For conventional Transformer~\cite{Vaswani2017}, $\mathbf{x}=\mathbf{h}^{N_E}_{1:m}$, which means the decoder layers only take the output of the final layer (\textit{i.e.} the $N_E$-th) as input.
This might omit some of the useful features from the lower encoder layers.
In this paper, given the outputs from all the encoder layers, $\{\mathbf{h}^{1:N_E}_{1:m} \}$, we take a weighted sum across all layers to obtain
the final image feature as:
\begin{equation}
\small
    \mathbf{x}_i=\alpha_l\sum_{l=1}^{N_E}\mathbf{h}^l_i
\end{equation}
where $\alpha_l$ are weights obtained using a softmax layer.

\subsection{Sequential Training with Inferred Labels}
\label{sec:training}

Training \name involves three steps: (i) training the Faster R-CNN object detector on Visual Genome dataset;
(ii) the trained Faster R-CNN is applied to train the encoder with Eq.~\ref{eq:sgg_llh} on Visual Genome dataset; (iii) when the encoder is well trained, it is further trained with the joint objective of scene graph generation and image captioning following Eq.~\ref{eq:total} on COCO dataset.
As the relation labels are not available on COCO dataset, we infer the labels for objects in COCO dataset using the encoder trained in step (ii).

The overall loss function in step (iii) is:
\begin{equation}
\small
    \mathcal{L} = \mathcal{L}_c + \lambda\mathcal{L}_r
\label{eq:total}
\end{equation}
with $\lambda$ being a hyper-parameter.

One possible variant of this training algorithm is to only use $\mathcal{L}_c$ without $\mathcal{L}_r$ as the training loss, as done in the literature work.
However, our Ablation studies in Tab.~\ref{tab:ablation} show that this variant cannot achieve as good results as those using the proposed training algorithm.
The main purpose of incorporating $\mathcal{L}_r$ is to ensure that the relational features learned do not vary much in step (iii) while being used to generate an accurate scene graph.

\section{Experiments}
\subsection{Datasets}

We evaluate \name on two large-scale publicly available datasets: MS COCO~\cite{Lin14} which is an image captioning dataset and Visual Genome~\cite{Krishna17} which is a scene graph generation dataset.

\textbf{COCO.} The dataset is the most popular benchmark for image captioning, which contains $82,783$ training images and $40,504$ validation images.
There are $5$ human annotated descriptions per image.
As the annotations of the official test set are not publicly available, we follow the widely used splits provided by~\cite{Karpathy2017}, where $5,000$ images are used for validation, $5,000$ for testing and the rest for training.
We convert all the descriptions in the training set to lower case and discard
rare words which occur less than 5 times, resulting in the final vocabulary with
$10,201$ unique words in the COCO dataset.

\textbf{Visual Genome.} The dataset is a large-scale image dataset for modeling the relationships between objects, which contains $108$K images with densely
annotated objects, attributes, and relations. 
To pre-train the Faster R-CNN object detector, we take $98$K for training, $5$K for validation and $5$K for testing. 
As part of images (about $50$K) in the Visual Genome are also found in COCO, the split of the Visual Genome is carefully selected to avoid contamination of the COCO validation
and test sets. 
We perform extensive cleaning and filtering of
training data, and train Faster R-CNN over the selected $1,600$ object classes. 
To pre-train the encoder of \name on the scene graph generation task, we adopt the same data split for training the object detector. 
Moreover, we select the top-$50$ frequent predicates in training data. 
The semantic relation detection model is thus trained over the $50$ relation classes plus a non-relation class.

\subsection{Methods \& Metrics}

We compare against three types of baselines.
(i) The CNN-LSTM~\cite{hochreiter1997long} based models: Up-Down~\cite{Anderson2018} which uses attention over regions of interest, NBT~\cite{Lu2018NeuralBT} that first generates a sentence `template' and then fill in by visual concepts identified by object detectors, Att2all~\cite{Rennie2017} that uses self-critical sequence training for image captioning, and AoA~\cite{huang2019attention} which uses attention on attention for encoding image regions and an LSTM language model;
(ii) Transformer-based models:
$\mathcal{M}^2$-T~\cite{cornia2020m2} which uses a mesh-like connectivity to learn prior knowledge, 
Image-T~\cite{He2020image}, an image transformer,
Object-T~\cite{Simao2019} that models the spatial relationship between objects, and
ETA~\cite{Li_2019_ICCV} which proposes the entangled attention mechanism;
(iii) The GCN-LSTM based models: VSUA~\cite{guo2019vsua} that uses GCNs to model the semantic and geometric interactions of the objects, GCN~\cite{Yao2018} which exploits pairwise relationships between image regions through a GCN, and SGAE~\cite{yang2019} which instead uses auto-encoding scene graphs.

For the caption generation evaluation, we follow the other baselines and use the BLEU-1 and BLEU-4 \cite{papineni2002}, ROUGE \cite{lin2004rouge}, METEOR \cite{denkowski2014meteor}, CIDEr \cite{Vedantam15cider} and SPICE~\cite{spice2016} metrics.

To evaluate the Scene Graph Generation, we divide it into three sub-tasks~\cite{zellers2018scenegraphs}: 
(i) Predicate Classification (PredCls), to predict $r_{i\rightarrow j}$ with $(b_i, o_i)$ and $(b_j, o_j)$ given;
(ii) Scene Graph Classification (SGCls), to predict the object labels $o_i$ and $o_j$ and the relationship $r_{i\rightarrow j}$ with $b_i$ and $b_j$ given;
(iii) Scene Graph Detection (SGDet), to directly predict $(b_i, o_i)$, $(b_j, o_j)$ and $r_{i\rightarrow j}$ from an image without the groundtruth bounding boxes or object labels.

The evaluation metrics we report are recall @$x$, where $x=20,50,100$.
Recall @$x$ computes the fraction of times the correct relationship is predicted in the top $x$ confident relationship predictions. 
It was first proposed in \cite{lu2016visual} and then widely adopted in other papers \cite{zellers2018scenegraphs,tang2018learning,tang2018learning}.
We notice that mean average precision (mAP) is another widely used metric. 
However, mAP is a pessimistic evaluation metric because we can not exhaustively annotate all possible relationships in an image. 
We thus do not report the results using this metric.
\subsubsection{Implementation Details}


\begin{table}[!t]
\footnotesize
\centering
\begin{tabular}{ccccccl}
\toprule
Method & B-1  & B-4 & M & R & C & S  \\
\midrule
Up-Down~\cite{Anderson2018} & 79.8 & 36.3 & 27.7 & 56.9 & 120.1 & 21.4 \\
Att2all~\cite{Rennie2017} & -- --  & 34.2 & 26.7 & 55.7 & 114.0 & -- --  \\
NBT~\cite{Lu2018NeuralBT}  & 75.5 & 34.7 & 27.1 & 54.7 & 107.2 & 20.1  \\
AoA~\cite{huang2019attention} & 80.2 & 38.9 & 29.2 & 58.8 & 129.8 & 22.4 \\
\midrule
ETA~\cite{Li_2019_ICCV} & 81.5 & 39.3 & 28.8 & 58.9 & 126.6 & 22.7 \\
Object-T~\cite{Simao2019} & 80.5 & 38.6 & 28.7 & 58.4 & 128.3 & 22.6 \\
Image-T~\cite{He2020image} & 80.8 & 39.5 & 29.1 & 59.0 & 130.8 & 22.8 \\
$\mathcal{M}^2$-T~\cite{cornia2020m2} & 80.8 & 39.1 & 29.2 & 58.6 & 131.2 & 22.6 \\
\midrule
GCN~\cite{Yao2018} & 80.5 & 38.2 & 28.5 & 58.3 & 127.6 & 22.0  \\
SGAE~\cite{yang2019} & 80.8 & 38.4 & 28.4 & 58.6 & 127.8 & 22.1  \\
VSUA~\cite{guo2019vsua} & -- -- & 38.4 & 28.5 & 58.4 & 128.6 & 22.0 \\
\midrule
\name & \textbf{82.3} & \textbf{39.8} & \textbf{29.7} & \textbf{59.8} & \textbf{131.9} & \textbf{23.0}  \\
\bottomrule
\end{tabular}
\caption{\small{Results on COCO dataset. We only report the single model results on the `Karpathy' test split. We highlight the \textbf{best} model in bold.}}
\label{tab:coco}
\end{table}

To represent image regions, we use Faster R-CNN with ResNet-101 finetuned on the Visual Genome dataset, thus obtaining a $2048$-dimensional feature vector for each region. 
To represent words, we use one-hot vectors and linearly project them to the input dimensionality of $512$. 
The dimension of the encoded bounding box, object label and the final image feature are $d_b=100$, $d_l=300$, and $d_h=512$ respectively.
We set the dimension of each layer to $d=512$ and the number of heads to $h=8$. 
We use the same number of encoders and decoders, thus having $N_E=N_D=3$.
We employ dropout with a probability $0.9$ after each attention and feed-forward layer. 
Training with the overall objective function (Eq.~\ref{eq:total}) is done following the learning rate scheduling strategy of~\cite{Vaswani2017} with a warmup equal to $10$K iterations. 
Then, during CIDEr optimization, we use a fixed learning rate of $5\times 10^{-6}$. 
We train all models using the Adam~\cite{kingma2015} optimizer.
We use Glove~\cite{pennington2014glove} embedding to initialize word embedding layer.
The total number of objects in one image varies from $10$ to $50$, depending on the IOU threshold that is set to $0.3$.
After some parameter tuning, we fix $\lambda=0.1$ at which \name provides the best CIDEr score.

\subsection{Evaluation}


\subsubsection{General Caption Generation}

We first evaluate our model with the general caption generation metrics.
We first compare the performances of our \name with those of several recent proposals for image captioning on the COCO `Karpathy' test split.
As shown in Tab.~\ref{tab:coco}, 
in general, the Transformer-based models outperforms other two types of baselines: the models using pre-trained CNN to encode image information and the LSTM decoder with attention to decode a caption and those using GCN to encode scene graph information and the LSTM decoder to generate a sentence.
This proves that Transformer can be used to better learn  high-level image features and is capable of decoding sentences with \rev{a} higher quality.
The GCN-based models outperform the CNN-based ones but \rev{perform worse than the ones using the Transformer}, which indicates that the scene graph information is useful for better learning the image features but still \rev{not well} explored.
Our proposed \name outperforms all other models. 
For example, it provides an improvement of $1.5$, $0.7$, $0.5$, $1.2$, $0.7$ and $0.4$ points over baseline model $\mathcal{M}^2$-T~\cite{cornia2020m2} on $6$ metrics respectively.
This demonstrates the effectiveness of concurrently exploiting the Transformer structure and scene graphs in extracting the relational image features.

We evaluate the model on the COCO online test server, composed of 40775 images for which annotations are not made publicly available.
The MSCOCO online testing results are
listed in Tab.~\ref{tab:coco_online}, our \name outperforms previous transformer based model on several evaluation metrics.

\begin{table*}[!t]
\centering
\begin{tabular}{ccccccccccl}
\toprule
\multirow{2}{*}{Method} & \multicolumn{2}{c|}{B-1}  & \multicolumn{2}{c|}{B-4} & \multicolumn{2}{c|}{M} & \multicolumn{2}{c|}{R} & \multicolumn{2}{c|}{C}   \\
\cmidrule{2-11}
&c5 & c40 &c5 & c40 &c5 & c40 &c5 & c40 &c5 & c40  \\
\midrule
Up-Down~\cite{Anderson2018} &  80.2 & 95.2 & 36.9 & 68.5 & 27.6 & 36.7 & 57.1 & 72.4 & 117.9 & 120.5 \\
Att2all~\cite{Rennie2017} &  78.1 & 93.7 & 35.2 & 64.5 & 27.0 & 35.5 & 56.3 & 70.7 & 114.7 & 116.7 \\
AoA~\cite{huang2019attention} &  81.0 & 95.0 & 39.4 & 71.2 & 29.1 & 38.5 & 58.9 & 74.5 & 126.9 & 129.6 \\
\midrule
ETA~\cite{Li_2019_ICCV} & 81.2 & 95.0 & 38.9 & 70.2 & 28.6 & 38.0 & 58.6 & 73.9 & 122.1 & 124.4  \\
Image-T~\cite{He2020image} & 81.2 & 95.4 & 39.6 & 71.5 & 29.1 & 38.4 & 59.2 & 74.5 & 127.4 & 129.6 \\
$\mathcal{M}^2$-T~\cite{cornia2020m2} & 81.6 & 96.0 & 39.7 & 72.8 & 29.4 & 39.0 & 59.2 & 74.8 & 129.3 & 132.1 \\
\midrule
GCN~\cite{Yao2018} & 80.8 & 95.9 & 38.7 & 69.7 & 28.5 & 37.6 & 58.5 & 73.4 & 125.3 & 126.5 \\
SGAE~\cite{yang2019} &  -- -- & -- -- & 37.8 & 68.7 & 28.1 & 37.0 & 58.2 & 73.1 & 122.7 & 125.5 \\
VSUA~\cite{guo2019vsua} & 79.9 & 94.7 & 37.4 & 68.3 & 28.2 & 37.1 & 57.9 & 72.8 & 123.1 & 125.5 \\
\midrule
\name & \textbf{82.0} & \textbf{96.7} & \textbf{40.1} & \textbf{73.2} & \textbf{29.8} & \textbf{39.5} & \textbf{59.9} & \textbf{75.2} & \textbf{129.9} & \textbf{132.8}  \\
\bottomrule
\end{tabular}
\caption{\small{Results on COCO dataset. We report the single model results on the COCO online test server. We highlight the \textbf{best} model in bold.}}
\label{tab:coco_online}
\end{table*}

\subsubsection{Ablation}
\begin{table}[!t]
\footnotesize
\centering
\begin{tabular}{cccccl}
\toprule
Method & B-4 & M & R & C & S  \\
\midrule
Trans.-2  & 35.7 & 27.4 & 56.4 & 121.3 & 20.5  \\
Trans.-3 & 36.5 & 27.8 & 57.0 & 123.6 & 21.1  \\
Trans.-4  & 36.3 & 27.6 & 56.5 & 121.5 & 20.8  \\
Trans.-5 & 36.1 & 27.5 & 56.8 & 121.9 & 20.7  \\
\midrule
Weighted Trans.  & 37.3 & 28.4 & 57.5 & 125.4 & 21.7  \\
\midrule
\name & \textbf{39.8} & \textbf{29.7} & \textbf{59.8} & \textbf{131.2} & \textbf{23.0}  \\
\midrule
\name $^{\ast}$ & 38.5 & 28.7 & 58.3 & 128.9 & 22.1  \\
\name $ - \mathcal{L}_r$ & 38.9 & 28.8 & 58.7 & 129.4 & 22.3  \\
\midrule
\name $-$ Weighted & 39.3 & 29.2 & 58.9 & 130.3 & 22.5  \\
\bottomrule
\end{tabular}
\caption{\small{Ablation study and comparison of \name variants. Results are reported on the `Karpathy' test split. $^{\ast}$ denotes that we fix the encoder during the caption training. We highlight the \textbf{best} model in bold.} }
\label{tab:ablation}
\end{table}

We first do ablation studies on the number of Transformer layers.
We start from the vanilla Transformer without using any other techniques proposed in the paper.
We vary the number of encoder and decoder layers from 2 to 6.
As shown in Tab.~\ref{tab:ablation}, the Transformer model with $3$ layers achieves the best results.
To evaluate the importance of keeping $\mathcal{L}_c$ in step (iii) of  Section~\ref{sec:training}, we conduct two ablation experiments.
We first keep the parameters of the encoder fixed, which means $\mathcal{L}_r$ is not used to update the parameters of the encoder.
We denote this variant as \name $^{\ast}$.
We find that the CIDEr score drops from $131.2$ to $128.9$.
We then remove $\mathcal{L}_r$ and only use $\mathcal{L}_c$ to update the parameters of the encoder and the decoder.
We denote this case as \name $-\mathcal{L}_r$.
We find that the CIDEr score drops as well.
We can thus conclude that using $\mathcal{L}_r$ helps to improve the quality of caption generation.


\subsubsection{Scene Graph Generation Evaluation}
\begin{table}[!t]
\footnotesize
\centering
\setlength\tabcolsep{1pt}
\begin{tabular}{cccccccccl}
\toprule
\multirow{2}{*}{Method} & \multicolumn{3}{c|}{\centering\textbf{SGGen}} & \multicolumn{3}{c|}{\centering\textbf{SGCls}} & \multicolumn{3}{c}{\centering\textbf{PredCls}} \\
\cmidrule{2-10}
 & R@20 & R@50 & R@100 & R@20 & R@50 & R@100 & R@20 & R@50 & R@100  \\
\midrule
IMP & 14.6 & 20.7 & 24.5 & 31.7 & 34.6 & 35.4 & 52.7 & 59.3 & 61.3  \\
MOTIFS  & 21.4 & 27.2 & 30.3 & 32.9 & 35.8 & 36.5 & 58.5 & 65.2 & 67.1  \\
VCTree & 22.0 & 27.9 & 31.3 & 35.2 & 38.1 & 38.8 & 60.1 & 66.4 & \textbf{68.1} \\
\name & \textbf{25.4} & \textbf{33.0} & \textbf{37.2} & \textbf{36.6} & \textbf{40.1} & \textbf{41.1} & \textbf{60.5}	& \textbf{66.7}	& \textbf{68.1}  \\
\bottomrule
\end{tabular}
\caption{\small{Scene graph generation evaluation results on Visual Genome dataset. 
We highlight the \textbf{best} model in bold.}}
\label{tab:sgg}
\end{table}

Generating scene graphs is essential not only because it provides a way of explaining the relationships between the objects in the image, but also the resource of the scene graphs when there is no other state-of-the-art scene graph generator available.
To evaluate the proposed \name in generating scene graphs, we compare it with other state-of-the-art scene graph generation methods.
IMP~\cite{xu2017scenegraph} solves the scene
graph inference problem using standard RNNs and learns to iteratively improves its predictions via message passing.
MOTIFS~\cite{zellers2018scenegraphs} is a stacked bi-directional LSTM architecture designed to capture higher order motifs in
scene graphs.
VCTree~\cite{tang2018learning} is a dynamic tree structure that places the objects in an image into a visual context which helps to improve the scene graph generation task.
The results are shown in Tab.~\ref{tab:sgg}.
\name achieves better results than all other three baselines on all the three tasks: \textbf{SGGen}, \textbf{SGCls} and \textbf{PredCls}.
This demonstrates the capability of our \name to exploit the relationships between the objects in the image.

\subsubsection{Qualitative Evaluation}

\begin{figure}
\small
    \centering
    \includegraphics[width=0.45\textwidth]{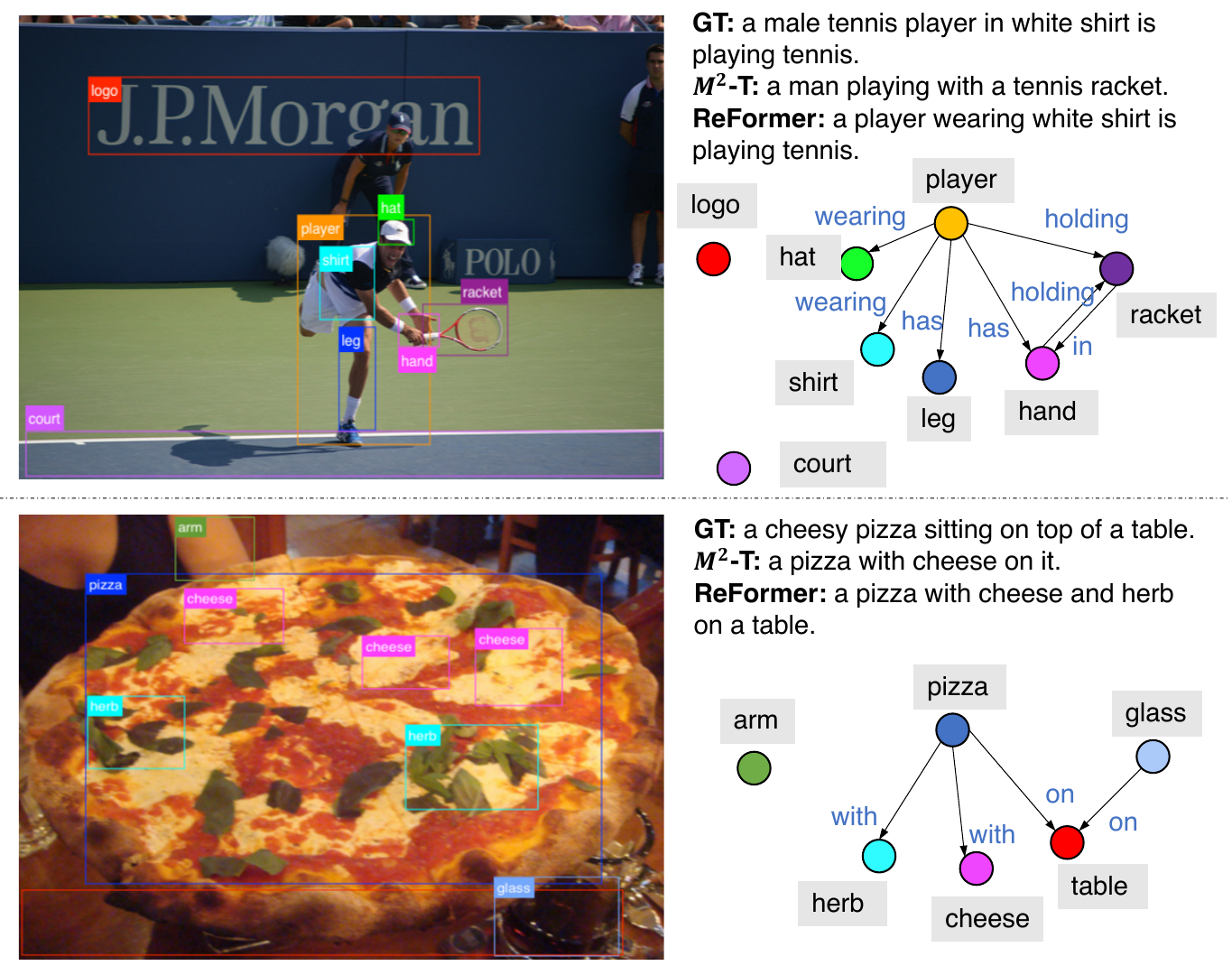}
    \caption{\small{Examples of captions generated by \name and the $\mathcal{M}^2$-T model, as well as the corresponding ground-truths. \name generates a scene graph to show the pair-wise relationships.}}
    \label{fig:sample}
\end{figure}

In Figure~\ref{fig:sample}, we show the image, groundtruth caption and the caption generated by $\mathcal{M}^2$ Transformer and \name.
Our model is able to not only generate a meaningful and more accurate caption than the baseline, but also generate a scene graph showing the relationships between the objects in the image.
With the graphs generated, captions become more expressive and explainable.
Interestingly, the graphs usually contain useful information to describe the image.
For instance, in the first example, the scene graph generated by \name tells us that the player is wearing a shirt and he is holding a racket.
While in the second example, the graph shows that there is a glass on the table and an arm from a person.
These information cannot be inferred from only the caption generated.
Since the average number of words in a sentence in the COCO dataset is $\approx 10$, it becomes very difficult for a captioner to describe details from the image using such short sentences.
Thus, a scene graph can be a good complementing component to the caption.

\section{Conclusion}


Exploring object relationships for image captioning is a challenging task, because it not only requires a strong encoder-decoder model to generate accurate captions but also a scheme to embed the relational information in the encoder. To effectively improve the image caption quality, we propose the use of \name to integrate the extraction of object relationship and caption generation into the same learning framework that the encoder 
can be more accurately trained. 
Our method achieves significant gains over COCO  dataset compared to the state-of-the-art models. However, there is still a room to improve the quality of image captioning. 
For example, to get more accurate scene graphs, one might use the online crowd-sourcing tools like Amazon Mechanical Turk to manually annotate the COCO dataset with relational labels. 

\section{Acknowledgements}
This work is supported in part by the National Science Foundation under Grants NSF 2134840 and NIH R01EB032218. We thank the reviewers for valuable discussions and feedbacks.

\bibliographystyle{ACM-Reference-Format}
\bibliography{egbib}

\appendix

\end{document}


\title{Supplementary Material for \\
ReFormer: The Relational Transformer for Image Captioning}

\author{First Author\\
Institution1\\
Institution1 address\\
{\tt\small firstauthor@i1.org}
\and
Second Author\\
Institution2\\
First line of institution2 address\\
{\tt\small secondauthor@i2.org}
}

\maketitle

\begin{abstract}
    This supplementary material provides more information about the Visual Genome dataset and the online evaluation on COCO test server.
\end{abstract}

\section{Dataset}
The Visual Genome dataset~\cite{Krishna17} consists of $51$ relation types, shown in Tab.~\ref{tab:vg_rel}.

\begin{table}[!p]
\caption{List of relation labels in Visual Genome dataset.}
\centering
\begin{tabular}{cp{8cm}cl}
\toprule
background, above, across, against, along, and, at,\\
attached to, behind, belonging to, between, carrying, \\
covered in, covering, eating, flying in, for, from, \\
growing on, hanging from, has, holding, in, in front of, \\
laying on, looking at, lying on, made of, mounted on, \\
near, of, on, on back of, over, painted on, parked on, \\
part of, playing, riding, says, sitting on, standing on, \\
to, under, using, walking in, walking on, watching, \\
wearing, wears, with \\
\bottomrule
\end{tabular}
\label{tab:vg_rel}
\end{table}

\begin{table*}[!t]
\centering
\begin{tabular}{ccccccccccl}
\toprule
\multirow{2}{*}{Method} & \multicolumn{2}{c|}{B-1}  & \multicolumn{2}{c|}{B-4} & \multicolumn{2}{c|}{M} & \multicolumn{2}{c|}{R} & \multicolumn{2}{c|}{C}   \\
\cmidrule{2-11}
&c5 & c40 &c5 & c40 &c5 & c40 &c5 & c40 &c5 & c40  \\
\midrule
Up-Down~\cite{Anderson2018} &  80.2 & 95.2 & 36.9 & 68.5 & 27.6 & 36.7 & 57.1 & 72.4 & 117.9 & 120.5 \\
Att2all~\cite{Rennie2017} &  78.1 & 93.7 & 35.2 & 64.5 & 27.0 & 35.5 & 56.3 & 70.7 & 114.7 & 116.7 \\
AoA~\cite{huang2019attention} &  81.0 & 95.0 & 39.4 & 71.2 & 29.1 & 38.5 & 58.9 & 74.5 & 126.9 & 129.6 \\
\midrule
ETA~\cite{Li_2019_ICCV} & 81.2 & 95.0 & 38.9 & 70.2 & 28.6 & 38.0 & 58.6 & 73.9 & 122.1 & 124.4  \\
Image-T~\cite{He2020image} & 81.2 & 95.4 & 39.6 & 71.5 & 29.1 & 38.4 & 59.2 & 74.5 & 127.4 & 129.6 \\
$\mathcal{M}^2$-T~\cite{cornia2020m2} & 81.6 & 96.0 & 39.7 & 72.8 & 29.4 & 39.0 & 59.2 & 74.8 & 129.3 & 132.1 \\
\midrule
GCN~\cite{Yao2018} & 80.8 & 95.9 & 38.7 & 69.7 & 28.5 & 37.6 & 58.5 & 73.4 & 125.3 & 126.5 \\
SGAE~\cite{yang2019} &  -- -- & -- -- & 37.8 & 68.7 & 28.1 & 37.0 & 58.2 & 73.1 & 122.7 & 125.5 \\
VSUA~\cite{guo2019vsua} & 79.9 & 94.7 & 37.4 & 68.3 & 28.2 & 37.1 & 57.9 & 72.8 & 123.1 & 125.5 \\
\midrule
\name & \textbf{82.0} & \textbf{96.7} & \textbf{40.1} & \textbf{73.2} & \textbf{29.8} & \textbf{39.5} & \textbf{59.9} & \textbf{75.2} & \textbf{129.9} & \textbf{132.8}  \\
\bottomrule
\end{tabular}
\caption{\small{Results on COCO dataset. We report the single model results on the COCO online test server. We highlight the \textbf{best} model in bold.}}
\label{tab:coco_online}
\end{table*}

\section{Extra Experiments}
We evaluate the model on the COCO online test server, composed of 40 775 images for which annotations are not made publicly available.
The MSCOCO online testing results are
listed in Tab.~\ref{tab:coco_online}, our \name outperforms previous transformer based model on several evaluation metrics.

{\small
\bibliographystyle{ieee_fullname}
\bibliography{egbib}
}